\documentclass[letterpaper, 10 pt, conference]{ieeeconf}  

\IEEEoverridecommandlockouts                              

\usepackage{amsmath} 
\usepackage{algorithm}
\usepackage{algpseudocode}
\usepackage{graphicx}
\usepackage{wrapfig}
\usepackage{booktabs}
\usepackage{tabularx}
\usepackage{amsfonts} 
\usepackage{url}
\usepackage{booktabs}
\usepackage{tabularx}
\usepackage{array}
\usepackage{xcolor, colortbl}

\title{\LARGE \bf
Urban Traffic Forecasting with Integrated Travel Time and Data Availability in a Conformal Graph Neural Network Framework
}

\author{Mayur Patil$^{1^*}$, Qadeer Ahmed$^{2}$ and Shawn Midlam-Mohler$^{3}$
\thanks{$^{1}$Graduate Student, Department of Mechanical and Aerospace Engineering, The Ohio State University 
        {\tt\small patil.151@buckeyemail.osu.edu}}%
\thanks{$^{2}$Assistant Professor, Mechanical and Aerospace Engineering, The Ohio State University 
        {\tt\small ahmed.358@osu.edu}}%
\thanks{$^{2}$Clinical Professor, Mechanical and Aerospace Engineering, The Ohio State University 
        {\tt\small midlam-mohler.1@osu.edu}}%
\thanks{*Corresponding Author}%
\thanks{This work was not supported by any organization}
}

\begin{document}

\maketitle
\thispagestyle{empty}
\pagestyle{empty}

\begin{abstract}

Traffic flow prediction is a big challenge for transportation authorities as it helps plan and develop better infrastructure. State-of-the-art models often struggle to consider the data in the best way possible, as well as intrinsic uncertainties and the actual physics of the traffic. In this study, we propose a novel framework to incorporate travel times between stations into a weighted adjacency matrix of a Graph Neural Network (GNN) architecture with information from traffic stations based on their data availability. To handle uncertainty, we utilized the Adaptive Conformal Prediction (ACP) method that adjusts prediction intervals based on real-time validation residuals. To validate our results, we model a microscopic traffic scenario and perform a Monte-Carlo simulation to get a travel time distribution for a Vehicle Under Test (VUT), and this distribution is compared against the real-world data. Experiments show that the proposed model outperformed the next-best model by approximately 24\% in MAE and 8\% in RMSE and validation showed the simulated travel time closely matches the 95th percentile of the observed travel time value.

\end{abstract}

\section{INTRODUCTION}

Traffic congestion in cities significantly impacts daily commutes, increases pollution, and makes urban life challenging. According to INRIX \cite{c1}, the average U.S. commuter spent 51 hours in traffic in 2022 — 15 hours more than in 2021 down from 99 hours in 2019. This persistent issue underscores the need for accurate traffic prediction. Predicting traffic conditions can aid in planning and avoiding congestion, yet traditional methods using historical data often fail to capture the complex dynamics of traffic effectively.

Lately, there has been significant progress in using machine learning techniques, particularly in understanding patterns in a spatio-temporal domain to predict traffic behavior. By viewing roads as a network of nodes and edges similar to a map, and utilizing models adept at handling sequential data such as how traffic changes throughout the day, we can make more informed decisions. Specifically, employing spatial models like Graph Neural Networks (GNNs) which are well-suited for mapping roads and intersections alongside models that handle the time-varying nature of traffic such as Recurrent Neural Networks (RNNs) can help address the complexities of traffic forecasting.

However, even with advancements in the models, it is still an arduous task to account for the actual physics, the associated uncertainties, and how the data is used in the best way possible. We leveraged Graph Convolution Network (GCN) with a Long Short-Term Memory (LSTM) architecture, inspired by Bing et al. \cite{c2}, in this study.  GCN is a type of GNN that applies the Convolutional Neural Network (CNN) concept in graph-structured data. It can perform a convolutional operation directly on the graph nodes and their neighboring nodes. Fundamentally, it can aggregate features of a selected node and its neighbors within an adjacency matrix.  We propose to use travel times between nodes and their neighbors (traffic stations) to make the graph model reflect a sense of realism. This helps us understand not just the layout of the roads but also how long it takes to get from one point to another. An LSTM architecture captures long-term dependencies in sequential data, using memory cells and gating mechanisms. This helps them to remember information over long sequences, addressing the vanishing gradient problem common in traditional RNNs.

We utilize data from Continuous Count Stations (CCS) and Non-Continuous Count Stations (N-CCS), combining high granularity data with sparse but insightful data. CCS provides detailed traffic counts every 15, 30, or 60 minutes year-round, while N-CCS offers limited data for short periods, adding sporadic yet valuable insights. To enhance predictions, we employ a weighting strategy that effectively balances the detailed CCS data with the sparse N-CCS data. Additionally, an adaptive conformal prediction (ACP) technique is employed to quantify uncertainty. To verify the effectiveness of our approach, validation is conducted by modeling a microscopic traffic scenario in SUMO (Simulation of Urban MObility) using predicted traffic flows and analyzing the VUT travel time distribution through Monte-Carlo simulation against real-world INRIX data.

The rest of this paper is organized as follows: Section 2 gives a comprehensive overview of the literature in this field. Section 3 discusses the methodology adopted followed by Section 4, describing the experiments and results. Finally, conclusions are presented in Section 5.

\section{Related Work}

\subsection{Spatio-Temporal Models}

Considerable work has gone into creating a specialized model in the subject of spatio-temporal traffic prediction that can handle both spatial and temporal dynamics for forecasting many different traffic states, including travel demand, trip duration, traffic speed, and traffic flow. In 2017, two prominent models were introduced: the Diffusion Convolutional Recurrent Neural Network (DCRNN) model \cite{c3} and the Spatio-Temporal Graph Convolutional Network (STGCN) model \cite{c2}. DCRNN used a diffusion process in conjunction with recurrent neural networks to control temporal dependencies, whereas STGCN integrated graph convolutions with gated temporal convolutions.

In 2019, the Temporal Graph Convolutional Network (T-GCN) was developed to improve the integration of spatial and temporal aspects in traffic forecasts by combining graph convolutions with gated recurrent units \cite{c4}. In order to better capture dynamic traffic patterns, attention mechanisms were concurrently added by the Attention-based Spatial-Temporal Graph Convolutional Networks (ASTGCN) \cite{c5}. The Graph Multi-Attention Network (GMAN) and Graph WaveNet were also developed around this time and they improved the model in terms of managing intricate spatial-temporal interactions by utilizing various attention processes \cite{c6}, \cite{c7}. Bi-component graph convolutions and the integration of graph convolution into LSTM cells were introduced by other models such as the Temporal Graph Convolutional Long Short-Term Memory (TGC-LSTM) network and the Multi-range Attentive Bicomponent Graph Convolutional Network (MFFB) \cite{c8}, \cite{c9}.  We also saw the introduction of the Spatio-Temporal Meta-learning framework (ST-MetaNet), applying meta-learning to spatio-temporal prediction, and the Dynamic Graph Convolutional Network (DGCN), which adjusted graph convolutional operations for constantly changing graphs \cite{c10}, \cite{c11}. Then came the Multi-Weighted Temporal Graph Convolutional network (MW-TGC) that refined spatial-temporal modeling even further by bringing in new ways to weigh temporal relationships and melding them with LSTM architectures \cite{c12}. Another study introduced the Structure Learning Convolution (SLC) framework \cite{c13} which bolstered the traditional CNN by incorporating structural information into the convolutional process.

In order to accommodate the dynamic nature of traffic, more recently, between 2021 and 2022, models like the Time-aware Multi-Scale Graph Convolutional Network (TmS-GCN) and the Attention-Accumulated Graph Recurrent Neural Network (AAGRNN) were developed by combining attention mechanisms and time-aware convolutions \cite{c14}, \cite{c15}.

These advancements show how spatio-temporal models may improve urban traffic management and planning, as each has made a distinct contribution to the evolution of traffic state prediction.

\subsection{Physics-Based Deep Learning Models}

While spatio-temporal models improve prediction efficacy, integrating physics-based principles adds realism. These models can capture fundamental relationships between traffic density, flow, and speed, reflecting the physical dynamics of traffic. Models like the Greenshields model \cite{c16} or the first-order Lighthill-Whitham-Richards (LWR) model provide insights into these interactions \cite{c17}, as shown in equation 1.

\begin{equation}
\frac{\partial \rho}{\partial t} + \frac{\partial (\rho v)}{\partial x} = 0
\end{equation}
where $\rho$ represents traffic density, $v$ is the speed of the traffic, $x$ is the spatial coordinate, and $t$ is time. Also, higher-order models like Payne–Whitham (PW) model \cite{c18} and Aw–Rascle–Zhang (ARZ) models \cite{c19} were developed to overcome the limitations intrinsic to LWR by introducing additional variables and considerations. The PW model is given by equations 2 and 3.
\begin{equation}
\begin{aligned}
\frac{\partial \rho}{\partial t} + \frac{\partial (\rho v)}{\partial x} & = 0, &
\frac{\partial v}{\partial t} + v\frac{\partial v}{\partial x} & = -\frac{1}{\rho}\frac{\partial P}{\partial x}
\end{aligned}
\end{equation}
where $P$ denotes the traffic pressure. The ARZ model differentiates between the actual and desired speeds of vehicles, which is reflected in its equations:
\begin{equation}
\begin{aligned}
\frac{\partial \rho}{\partial t} + \frac{\partial (\rho v)}{\partial x} & = 0, &
\frac{\partial y}{\partial t} + v\frac{\partial y}{\partial x} & = 0
\end{aligned}
\end{equation}
where $y$ represents the difference between the desired and actual speed. One of the advancements was shown when microscopic traffic models like the Intelligent Driver Model (IDM) were developed \cite{c31}. This model is used in SUMO to dictate the behavior of individual vehicles within the simulation:
\begin{equation}
\frac{dv}{dt} = a \left[ 1 - \left(\frac{v}{v_{\text{desired}}}\right)^4 - \left(\frac{s^*(v, \Delta v)}{s}\right)^2 \right]
\end{equation}

\begin{equation}
s^*(v, \Delta v) = s_0 + \max\left(0, vT + \frac{v\Delta v}{2\sqrt{a b}}\right)
\end{equation}
where \( v \) is the vehicle's current speed, \( v_{\text{desired}} \) is its desired speed, \( \Delta v \) is the speed difference with the vehicle ahead, \( s \) is the current gap, \( s^* \) is the desired safety distance, \( s_0 \) is the minimum static gap, \( T \) is the time headway, \( a \) is the maximum acceleration, and \( b \) is the comfortable braking deceleration.
SUMO uses IDM to generate vehicle routes based on edge count data to simulate realistic vehicle interactions within the network. Later studies like Physics-Informed Neural Networks (PINNs) utilized the LWR model in the presence of sparse data grids and proved their efficacy \cite{c20}. Additionally, the Spatio-Temporal Dynamic Network (STDEN) model utilized potential energy fields in order to take into account the dynamic impact of adjacent regions, consequently improving forecasting capabilities \cite{c21}. 

While the above-discussed investigations improve accuracy, they introduced an extra computational expense by incorporating physics-based parameters into the neural network training loss functions. To address this, we propose adding the travel/transit times between the traffic stations to the adjacency matrix to reduce computational overhead while maintaining accuracy. This intuitively makes sense as it is directly related to the fundamental parameters- speed, density, and flow.

\subsection{Uncertainty Quantification}

Traffic flow prediction comes with many unpredictable elements like sudden accidents, changing weather, and the number of cars on the road at any time. The traditional ways to quantify uncertainties use Bayesian inference, quantile regression, and other ensemble-based models \cite{c22}, \cite{c23}, and \cite{c24}. There is a comparatively newer technique called Conformal Prediction (CP) that has gotten a lot of attention as it comes with a coverage guarantee of the actual outcomes and it does not depend on a well-calibrated model. It also yields a transparent and traceable way of obtaining confidence intervals \cite{c25}.  There is an even more advanced version of CP called Adaptive Conformal Prediction (ACP) \cite{c26}, which changes the prediction ranges as new data comes in. It keeps on updating these ranges to make sure they stay accurate even when the incoming data changes. However, we modified this to a simpler form by not considering the additional factor $\gamma$ (learning rate) and directly used the residuals (differences in the past prediction values) to maintain the desired confidence level.


\section{Methodology}
\subsection{Graph Network}
The road network is represented as a graph $\mathcal{G} = (\mathcal{V}, \mathcal{E})$, where $\mathcal{V}$ denotes the set of nodes that are traffic count stations and $\mathcal{E}$ represents the set of edges that connects these nodes. Each node $v_i \in \mathcal{V}$ has traffic flow measurements, and each edge $e_{ij} \in \mathcal{E}$ shows the road stretch between two nodes $v_i$ and $v_j$. Here, $i$ and $j$ are the indices representing nodes in the traffic network. The adjacency matrix $\mathbf{A} \in \mathbb{R}^{|\mathcal{V}| \times |\mathcal{V}|}$ is an important component in a graph network, that contains the information regarding connectivity and interaction between nodes. We introduced $\mathbf{A}$ based on travel times derived from average speeds and distances between the traffic stations, alongside data availability scores for CCS and N-CCS stations. Let us assume, $C_i$ denotes the number of counts at N-CCS$_i$. The data availability for N-CCS stations can formulated by score normalization as:

\begin{equation}
A_i^{N-CCS} = \frac{C_i}{\max_{j} (C_j)}
\end{equation}
where $C_i$ denotes counts at station $N-CCS_i$ and $\max_{j} (C_j)$ is the maximum count across all N-CCSs. The CCS stations are assumed fully available, denoted by $\mathbf{A}^{\text{CCS}} = \mathbf{1}_{|\mathcal{V}_{\text{CCS}}|}$, where $\mathcal{V}_{\text{CCS}}$ is the set of CCS nodes. The travel times $\mathbf{T}$, are computed as: 
\begin{equation}
T_{ij} = \frac{D_{\text{combined}_{ij}}}{S_{ij}}
\end{equation}
where $D_{\text{combined}_{ij}}$ is the distance between the nodes, and $S_{ij}$ is the average speed on the edge connecting these nodes. We use these travel times to adjust the adjacency matrix, favoring shorter travel times, specifically:
\begin{equation}
f(T_{ij}) = \frac{1}{T_{ij} + \epsilon}
\end{equation}
where $\epsilon$ is a small constant to ensure non-zero denominators. 

The combined adjacency matrix $\mathbf{A}_{\text{combined}}$ is initialized by applying a Gaussian kernel \cite{c2} to the normalized travel times:
\begin{equation}
\mathbf{A}_{\text{combined}} = \exp\left(-\frac{\mathbf{T^2_{norm}}}{2\sigma^2}\right)
\end{equation}
where $\sigma^2$ is the variance parameter for the Gaussian kernel, controlling the spread of the weights in the adjacency matrix.

At this point transformation can be applied to $\mathbf{A}_{\text{combined}}$ using 8:
\begin{equation}
\mathbf{A}_{\text{combined}[i,j]} = f(T_{ij})
\end{equation}
Then, $\mathbf{A}_{\text{combined}}$ is operated with the availability scores for both CCS and N-CCS stations to form the final $\mathbf{A}_{\text{modified}}$ matrix:
\begin{equation}
\mathbf{A}_{\text{modified}[i, j]} = \mathbf{A}_{\text{combined}[i, j]} \cdot A_i \cdot A_j
\end{equation}
where $A_i$ and $A_j$ are the availability scores for nodes $i$ and $j$ respectively. This formulation makes the weighted adjacency matrix $\mathbf{A}_{\text{modified}}$ to include actual travel times and data availability to further be considered in the GCN model for spatial connections.

\subsection{Graph Convolution Network (GCN)}
In our model, we used GCN \cite{c2} to understand spatial interconnections of the road network. In a GCN, nodes represent traffic stations, and edges signify the connecting roads. Each node gets a set of features and also information about its neighboring nodes. We update this combined information in steps. In each step, every point looks at the information of the points it is directly connected to and gathers up this information known as a message-passing mechanism:

\begin{equation}
    \mathbf{m}_i^{(l+1)} = \text{A}^{(l)}\left(\left\{ \mathbf{h}_j^{(l)} : j \in \mathcal{N}(i) \right\}\right)
\end{equation}

\begin{equation}
    \mathbf{h}_i^{(l+1)} = \text{U}^{(l)}\left(\mathbf{h}_i^{(l)}, \mathbf{m}_i^{(l+1)}\right)
\end{equation}

where ${m}_i^{(l+1)}$ represents the aggregated message for node $i$ at layer $l+1$, $\text{A}^{(l)}(\cdot)$ combines the information of neighboring nodes ${h}_j^{(l)}$, $j$ is in the set of neighbors $N(i)$ of node $i$, ${h}_i^{(l)}$ denotes the information of node $i$ at layer $l$, and $\text{U}^{(l)}(\cdot)$ is an update function generating new information for node $i$ based on its previous embedding and the aggregated message. The network then updates the node embeddings using a combination of the aggregated messages and the nodes' previous embeddings, often through a transformation that involves learnable parameters and non-linear activation functions.

\subsection{LSTM with Attention Mechanism}
We used an LSTM model to capture temporal traffic patterns such as daily road congestion. LSTM processes GCN data sequentially but to emphasize on important moments, we added an attention mechanism \cite{c27}. This mechanism assigns weights to each time step, enhancing the final prediction by focusing on the most critical data points.
\begin{equation}
    \mathbf{c} = \sum_{t=1}^{T} \alpha_t \mathbf{h}_t
\end{equation}
\begin{equation}
    \alpha_t = \frac{\exp(e_t)}{\sum_{k=1}^{T} \exp(e_k)}
\end{equation}
\begin{equation}
    e_t = \tanh(\mathbf{h}_t \mathbf{W}_{att} + \mathbf{b}_{att})
\end{equation}
where $\mathbf{h}_t$ is the LSTM output at time step $t$, $\mathbf{W}_{att}$ and $\mathbf{b}_{att}$ are the trainable parameters of the attention layer, and $\mathbf{c}$ is the context vector that serves as the input to the subsequent dense layer for final prediction.

\subsection{Adaptive Conformal Prediction (ACP)}
We used a modified CP framework called ACP to quantify the uncertainty in the predictions. ACP provides prediction intervals that encompass actual traffic flow values and adjusts these intervals based on observed residuals \cite{c26}. Residuals between actual and predicted values are calculated, defining the quantile value:
\begin{equation}
    PI_{\hat{\mathbf{F}}} = [\hat{\mathbf{F}} - q_{\text{adjusted}}, \hat{\mathbf{F}} + q_{\text{adjusted}}]
\end{equation}
where $\hat{F}$ is the predicted traffic flow, and $q_{\text{adjusted}}$ is the quantile of the absolute residuals observed on the validation set, corresponding to a significance level $\alpha$ that represents the proportion of future predictions we expect to fall inside the prediction intervals. The adaptive mechanism updates $q_{\text{adjusted}}$ at the end of each training epoch based on the latest residuals. A callback function then evaluates the performance on the validation set, recalculates $q_{\text{adjusted}}$, and adjusts the prediction intervals accordingly. We used $\alpha=0.1$, meaning we expected $90\%$ of the times the true values will fall within the predicted interval.

\subsection{Optimization}
In training the model, we used Mean Squared Error (MSE) as the loss function.
\begin{equation}
    \mathcal{L} = \frac{1}{N} \sum_{i=1}^{N} (\hat{y}_i - y_i)^2
\end{equation}
where $N$ is the number of samples, $\hat{y}_i$ is the predicted traffic flow, and $y_i$ is the actual traffic flow. We used the RMSprop optimization algorithm with a learning rate of 0.0002 and is dynamically adjustable for each parameter to improve convergence speed and handle non-stationary objectives. To prevent overfitting, we implemented an early stopping technique that halts training if no improvement is seen over a certain epoch. Additionally, a custom callback function updates the ACP quantile based on validation dataset residuals ensuring accurate prediction ranges. 

The overall model architecture showing interactions between its components (GCN layer for spatial data, an LSTM with Attention for temporal patterns, and an ACP layer for uncertainty quantification) is illustrated in Figure 1.

\begin{figure}[htbp]
\centering
\includegraphics[width=\linewidth]{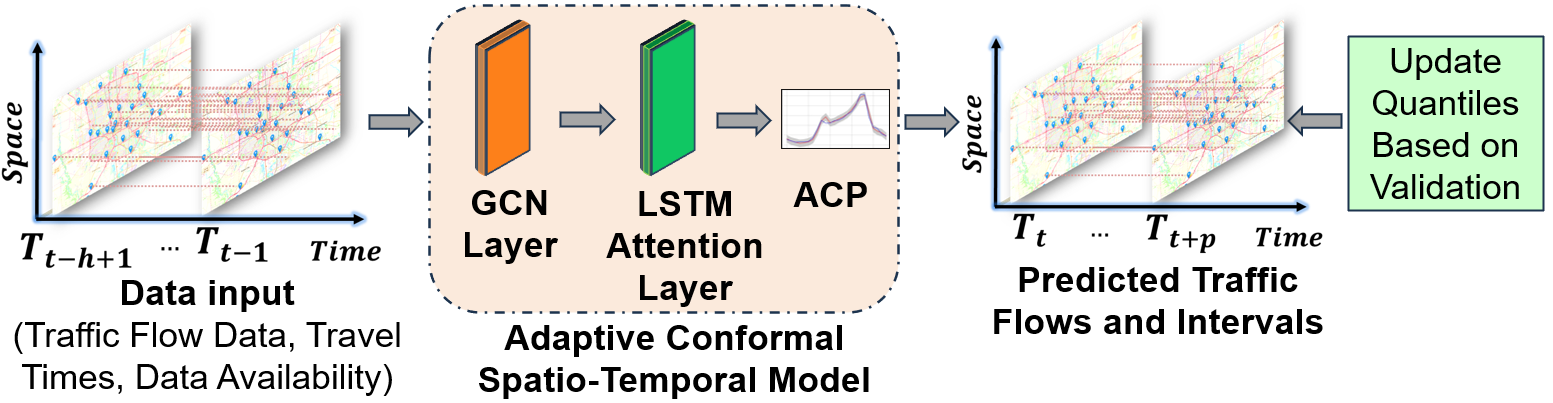}
\caption{Overall Model Architecture}
\label{fig:my_label}
\end{figure}

The pseudo-code summarizing the approach is outlined in Algorithm 1.\\

\begin{algorithm}
\caption{Traffic Flow Prediction with GCN-LSTM-ACP}
\begin{algorithmic}[1]
\State \textbf{Input:} Traffic flow data $\mathbf{F}$, route travel times $\mathbf{T}$, availability scores $\mathbf{A}$
\State \textbf{Output:} Predicted traffic flow $\hat{\mathbf{F}}$, prediction intervals $\mathbf{PI}$

\Procedure{AdjMat}{$\mathbf{T}, \mathbf{A}$}
    \State Normalize $\mathbf{T}$ to $\mathbf{T}_{\text{norm}}$ using $\max(\mathbf{T})$
    \State $\mathbf{A}_{\text{modified}} = \exp\left(-\frac{\mathbf{T}^2_{norm}}{2\sigma^2}\right) \odot (\mathbf{A} \mathbf{A}^T)$
\EndProcedure

\Procedure{GraphConv}{$\mathbf{F}, \mathbf{A}_{\text{modified}}$}
    \State $\mathbf{h}_i^{(l+1)} = \text{U}^{(l)}\left(\mathbf{h}_i^{(l)}, \sum_{j \in \mathcal{N}(i)} \mathbf{A}_{\text{modified}}[i, j] \cdot \mathbf{h}_j^{(l)}\right)$
\EndProcedure

\Procedure{LSTM-Att}{$\mathbf{H}$}
    \State $\alpha_t = \text{softmax}(\tanh(\mathbf{H} \mathbf{W}_{\text{att}} + \mathbf{b}_{\text{att}}))$
    \State $\mathbf{c} = \sum (\alpha_t \cdot \mathbf{H})$
    \State $\hat{\mathbf{F}} = \text{Dense}(\mathbf{c})$
\EndProcedure

\Procedure{TrainAndUpdate}{$\mathbf{F}_{\text{train}}, \mathbf{F}_{\text{val}}$}
    \State Update $q_{\text{adjusted}}$ from $|\hat{\mathbf{F}} - \mathbf{F}_{\text{val}}|$ during training
\EndProcedure

\Procedure{ACP}{$\hat{\mathbf{F}}, \alpha$}
    \State $q_{\text{adjusted}} = \text{quantile}(|\hat{\mathbf{F}} - \mathbf{F}_{\text{val}}|, 1 - \alpha)$
    \State $\mathbf{PI} = [\hat{\mathbf{F}} - q_{\text{adjusted}}, \hat{\mathbf{F}} + q_{\text{adjusted}}]$
\EndProcedure

\Procedure{Predict}{$\mathbf{F}_{\text{test}}$}
    \State $\hat{\mathbf{F}}_{\text{test}} = \text{Model Predict}(\mathbf{F}_{\text{test}})$
    \State \textbf{call} \textsc{ACP}$(\hat{\mathbf{F}}_{\text{test}}, 0.1)$
\EndProcedure

\State \textbf{return} $\hat{\mathbf{F}}_{\text{test}}, \mathbf{PI}$
\end{algorithmic}
\end{algorithm}

The techniques employed i.e. batch processing, early stopping, and RMSprop optimization, ensure computational efficiency and scalability. It also expedites training and enables the effective handling of large datasets, making the system suitable for real-time traffic prediction.

\section{Experiment}
\subsection{Dataset}
We used the Ohio ODOT-TCDS data subset with traffic information from 51,777 N-CCS and 396 CCS nodes, recorded every 15 minutes in 2019. Because the trip times between stations are directional, the adjacency
matrix in Figure 2 is asymmetrical. For example, node 1 connects to nodes 2-9 (1-8 minutes), while the reverse direction takes much longer.

\begin{figure}[htbp]
\centering
\includegraphics[width=\linewidth]{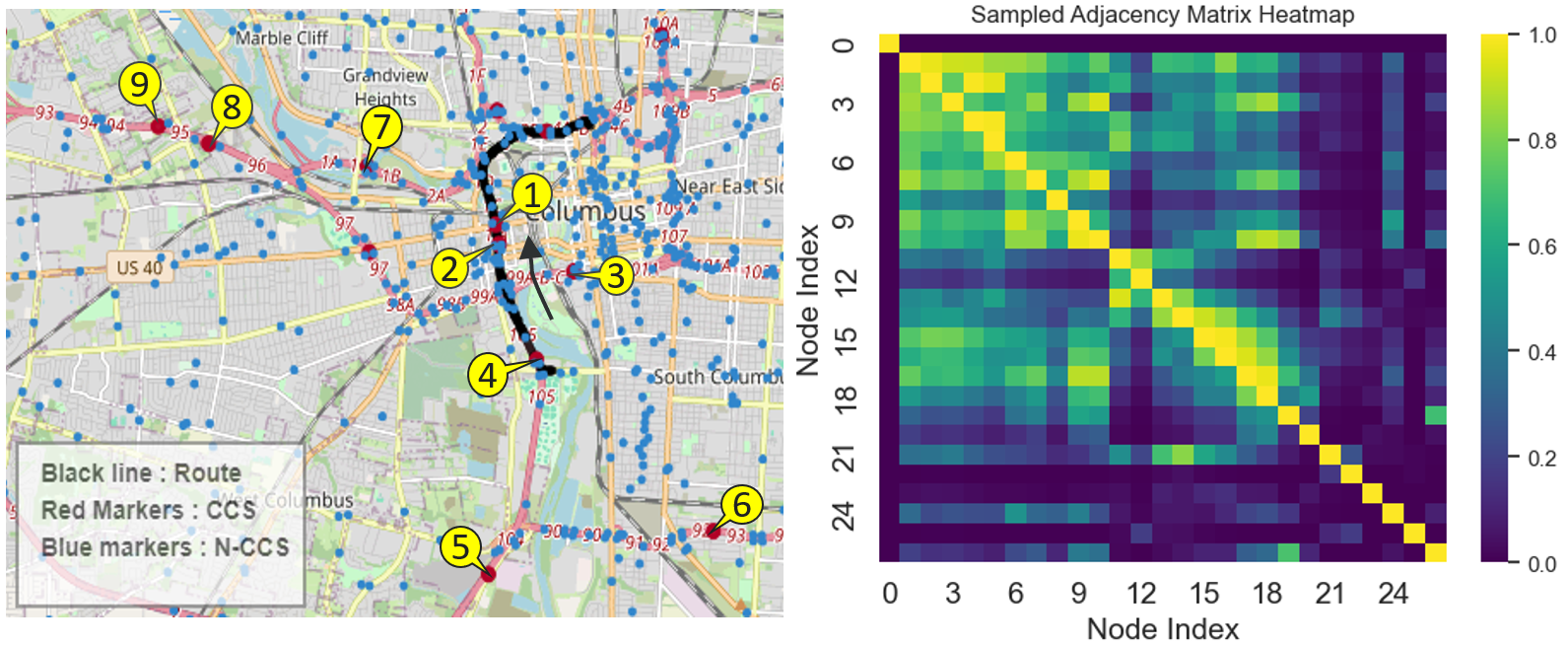}
\caption{Sampled Route (left); Heat Map of Adjacency Matrix (right)}
\label{fig:my_label}
\end{figure}

\subsection{Evaluation Metric and Baselines}
Mean Absolute Errors (MAE) and Root Mean Squared Errors (RMSE) are adopted to measure and evaluate the performance of different methods. 

\begin{equation}
\text{MAE} = \frac{1}{N} \sum_{i=1}^{N} |\mathbf{F}_i - \hat{\mathbf{F}}_i|
\end{equation}

\begin{equation}
\text{RMSE} = \sqrt{\frac{1}{N} \sum_{i=1}^{N} (\mathbf{F}_i - \hat{\mathbf{F}}_i)^2}
\end{equation}

We compare our method with the following baselines: 1) Spatio-Temporal Graph Convolutional Networks (STGCN) \cite{c2}; 2) Auto-Regressive Integrated Moving Average (ARIMA) [28]; 3) Historical Average (HA); 4) Feed-Forward Neural Network (FNN); and 5) LSTM.


We also report mean prediction interval width (MPIW) and prediction interval coverage probability (PICP) to quantify uncertainty.

\begin{equation}
\text{PICP} = \frac{1}{N} \sum_{i=1}^{N} \mathbf{1}(\hat{\mathbf{F}}_i^L \leq \mathbf{F}_i \leq \hat{\mathbf{F}}_i^U)
\end{equation}

\begin{equation}
\text{MPIW} = \frac{1}{N} \sum_{i=1}^{N} (\hat{\mathbf{F}}_i^U - \hat{\mathbf{F}}_i^L)
\end{equation}
where \(\hat{\mathbf{F}}_i^L\) and \(\hat{\mathbf{F}}_i^U\) are the lower and upper bounds of the prediction interval for the \(i\)-th observation, respectively.

\subsection{Results}
Table I demonstrates the results for our model and comparison with the baselines on the ODOT-TCDS dataset. Our proposed model shows a reduction of 24\% in MAE and 8\% in RMSE compared with STGCN. We used a forecast horizon of 15 min with a 24 hours of look-back period. 


\definecolor{headercolor}{gray}{0.9}

\begin{table}[h]
\caption{Comparison of MAE and RMSE across different models.}
\label{table_performance_comparison}
\centering
\begin{tabularx}{\columnwidth}{|l|>{\centering\arraybackslash}X|>{\centering\arraybackslash}X|}
\hline
\rowcolor{headercolor} \textbf{Method} & \multicolumn{2}{c|}{ODOT-TCDS sub-sample (15 min)} \\
\cline{2-3} 
\rowcolor{headercolor} & \textbf{MAE} & \textbf{RMSE} \\
\hline
STGCN & 0.29 & 0.38 \\
\hline
ARIMA & 0.67 & 1.08 \\
\hline
HA & 0.51 & 0.75 \\
\hline
FNN & 0.39 & 0.53 \\
\hline
LSTM & 0.34 & 0.49 \\
\hline
Proposed Model & \textbf{0.22} & \textbf{0.35} \\
\hline
\end{tabularx}
\end{table}

It can be observed that ARIMA and HA perform poorly due to their inability to capture non-linear and dynamic traffic patterns. In contrast, FNN and LSTM improve accuracy by modeling non-linear relationships and temporal dependencies. A much better model, STGCN, further enhances predictions by incorporating spatial features. However, our proposed model surpasses STGCN by integrating travel times and data availability, more effectively capturing complex urban traffic dynamics.

Figure 3 illustrates the loss and MAE over epochs for the training and validation sets. The plots suggest that our model can achieve an acceptable convergence. 
\begin{figure}[htbp]
\centering
\includegraphics[width=0.8\linewidth, height=0.15\textheight, keepaspectratio]{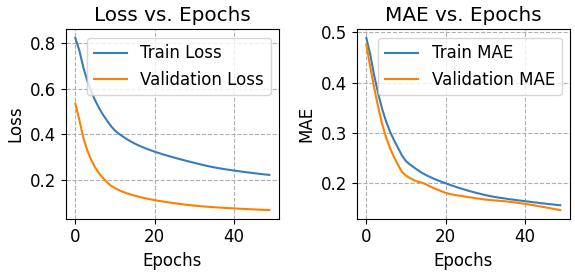}
\caption{Loss versus Epochs (left); MAE versus Epochs (right)}
\label{fig:my_label}
\end{figure}
The results for uncertainty quantification for a selected CCS node are shown in Figure 4.

\begin{figure}[htbp]
\centering
\includegraphics[width=\linewidth]{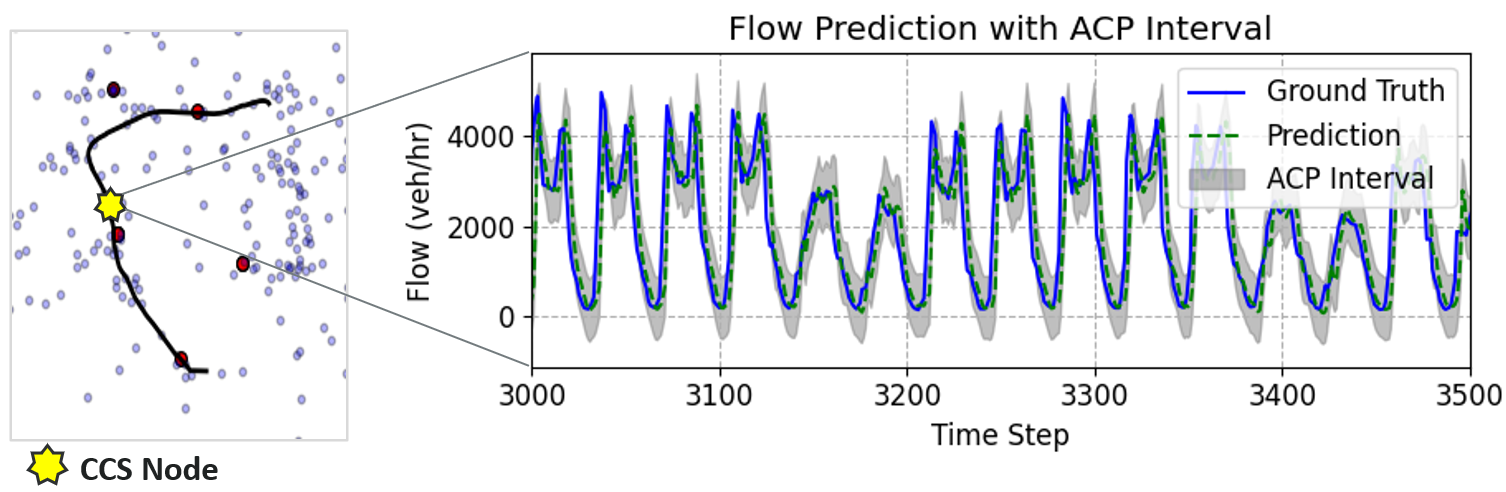}
\caption{Prediction with Uncertainty Bounds}
\label{fig:my_label}
\end{figure}

The proposed methodology for quantifying uncertainty showed a PICP of 90.02\% and an MPIW of 1.03 which conveys that the resulting coverage probability falls within the expectation of 90\% coverage rate as considered in ACP.

\subsection{Traffic Model Calibration and Validation}
We utilized the \textit{routesampler} tool in SUMO following methodology in \cite{c30} to generate traffic elements and origin-destination pairs based on vehicle counts. This tool uses linear programming to match real-world traffic counts, adjusting parameters like vehicle departure intervals, traffic volume, driver behavior, and vehicle dynamics for realistic simulations. We extracted the ACP's upper bound predictions to mimic peak-hour traffic from 7:30 to 8:30 am on weekdays. Then, a Monte-Carlo simulation was conducted by varying the departure time of the generated vehicles for 200 runs and measured the VUT travel time parameter to evaluate the model's performance which was compared against the INRIX probe data \cite{c32}, as shown in Figure 5.

\begin{figure}[htbp]
\centering
\includegraphics[width=\linewidth]{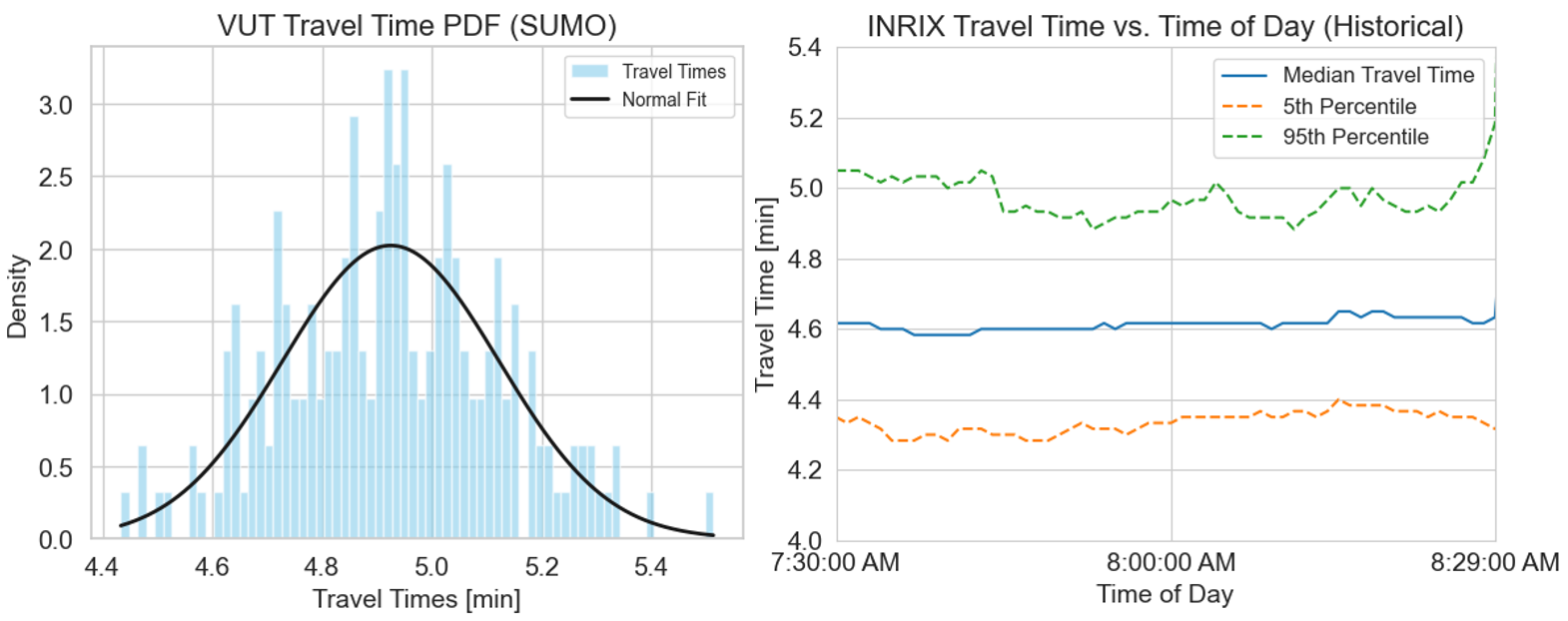}
\caption{PDF of VUT travel times from Monte-Carlo SUMO simulations (left); INRIX Probe Data (right) during 7:30AM - 8:30AM }
\label{fig:my_label}
\end{figure}

While discrepancies were present, as is expected in any model trying to mimic complex real-world phenomena, the overall trend of travel times during the selected period was captured reasonably well by the simulation. The distribution mean is around 4.93 minutes which is close to the 95th percentile from the INRIX probe data.

\section{CONCLUSIONS}


In this study, we propose an approach that uses travel times to define the weights in a GNN’s adjacency matrix, simplifying it while maintaining a physics-like perspective. We also include data from all types of traffic monitoring stations for comprehensive coverage. To address uncertainty, we apply the ACP method for better reliability. It outperforms existing methods by effectively integrating spatial and temporal data, capturing long-term dependencies, and providing dynamic uncertainty bounds. However, its complexity and data requirements may limit use in areas with sparse data. Finally, we validate the predictions through Monte-Carlo simulation in SUMO, assessing VUT travel time distribution against INRIX probe data, which showed that the predicted travel times closely matched the real-world data.

To improve the model’s forecasting ability, future directions for this work will include a variety of urban contexts and the addition of outside variables like traffic accidents and weather fluctuations.

\addtolength{\textheight}{-12cm}   








\end{document}